\documentclass[10pt,twocolumn,letterpaper]{article}

\usepackage{cvpr}      
\definecolor{cvprblue}{rgb}{0.21,0.49,0.74}
\usepackage[pagebackref,breaklinks,colorlinks,allcolors=cvprblue]{hyperref}

\usepackage{tabularx}
\usepackage{booktabs}
\usepackage{subcaption}
\usepackage{graphicx}
\usepackage[ruled,vlined]{algorithm2e}
\usepackage{bbm}
 
\usepackage[accsupp]{axessibility} 
\def\paperID{AI4RWC-23} 
\def\confName{CVPRW}
\def\confYear{2026}

\title{Event-Level Detection of Surgical Instrument Handovers in Videos with Interpretable Vision Models}

\author{
Katerina Katsarou$^{1*}$\thanks{Equal contribution.}\thanks{Corresponding author: email \newline \texttt{aikaterini.katsarou@hhi.fraunhofer.de}}\thanks{This research was supported by the German Federal Ministry of Research, Technology and Space (BMFTR) with project Kiara [grant: 16SV9036].}\quad
George Zountsas$^{12}\footnotemark[1]$\quad
Karam Tomotaki-Dawoud$^{1}$\quad
Alexander Ehrenhoefer$^{12}$\quad
\and
Paul Chojecki$^{1}$\quad
David Przewozny$^{1}$\quad
Detlef Runde$^{1}$\quad
Igor Maximilian Sauer$^{3}$\quad
\and
Amira Mouakher$^{4}$\quad
Sebastian Bosse$^{1}$
\\[6pt]
$^{1}$Fraunhofer HHI, Berlin, Germany\quad
$^{2}$Technical University of Berlin, Germany\\[4pt]
$^{3}$Charité - Universitätsmedizin Berlin, Germany\quad
$^{4}$Université de Perpignan, Perpignan, France
}

\begin{document}
\maketitle
\maketitle
\footnotetext{The source code is available on our \href{https://gitlab.hhi.fraunhofer.de/dawoud/surgical-shield.git}{Git}.}

\begin{abstract}
Reliable monitoring of surgical instrument exchanges is essential for maintaining procedural efficiency and patient safety in the operating room. Automatic detection of instrument handovers in intraoperative video remains challenging due to frequent occlusions, background clutter, and the temporally evolving nature of interaction events. We propose a spatiotemporal vision framework for event-level detection and direction classification of surgical instrument handovers in surgical videos. The model combines a Vision Transformer (ViT) backbone for spatial feature extraction with a unidirectional Long Short-Term Memory (LSTM) network for temporal aggregation. A unified multi-task formulation jointly predicts handover occurrence and interaction direction, enabling consistent modeling of transfer dynamics while avoiding error propagation typical of cascaded pipelines. Predicted confidence scores form a temporal signal over the video, from which discrete handover events are identified via peak detection. Experiments on a dataset of kidney transplant procedures demonstrate strong performance, achieving an F1-score of 0.84 for handover detection and a mean F1-score of 0.72 for direction classification, outperforming both a single-task variant and a VideoMamba-based baseline for direction prediction while maintaining comparable detection performance. To improve interpretability, we employ Layer-CAM attribution to visualize spatial regions driving model decisions, highlighting hand–instrument interaction cues.
\end{abstract}    
\section{Introduction}
\label{sec:intro}

The operating room represents a high-stakes clinical environment in which procedural efficiency and patient safety must be simultaneously maintained. 
Among the factors influencing surgical workflow quality is the seamless exchange of instruments between the surgeon and the assistant. 
Instrument handovers are fundamental to preserving procedural flow, reducing interruptions, and sustaining cognitive continuity during surgery. 
Disruptions or inefficiencies during these interactions may contribute to workflow interruptions, prolonged procedure times, and increased risks of error.
Beyond workflow efficiency, reliable instrument exchange is closely linked to patient safety. 
Failures in instrument countability remain a contributing factor to adverse events such as retained surgical instruments (RSIs), 
a preventable yet clinically significant complication associated with substantial medical and economic consequences~\cite{kooijmans2024surgical}. 
Reliable monitoring of instrument transfers is therefore essential not only for optimizing surgical workflows but also for supporting procedural safety. 
These challenges motivate the development of automated systems capable of analyzing surgical interactions directly from intraoperative video data.

Recent advances in artificial intelligence and computer vision have enabled vision-based analysis of surgical environments, including instrument tracking, action recognition, and workflow understanding. 
Transformer-based architectures have demonstrated strong representation learning capabilities by modeling complex spatial dependencies. 
Prior work has explored the integration of large-scale pretrained visual models for surgical scene understanding. For instance, the SurgiGuard framework \cite{SurgiGuard} leveraged CLIP-based visual representations combined with graph-based reasoning 
to detect surgical instrument handovers and model relationships between surgical entities. 
By combining vision-based representations with knowledge graphs, SurgiGuard demonstrated the feasibility of automated interaction detection and visualization.

Despite these advances, automatic handover detection remains highly challenging in real clinical environments. 
Intraoperative videos acquired during live surgical procedures exhibit severe occlusions, cluttered backgrounds, dynamic illumination, 
and substantial viewpoint variability.
Moreover, handover interactions are inherently temporal phenomena unfolding across multiple frames, 
rendering frame-level representations alone is insufficient for capturing transfer dynamics. 
These characteristics motivate spatiotemporal architectures capable of jointly modeling spatial structure and temporal evolution.

In this work, we propose a framework for surgical instrument handover analysis using intraoperative videos collected from real surgical procedures. 
Unlike approaches evaluated on simulated or laboratory-controlled recordings, the proposed approach is designed for unconstrained clinical environments characterized by visual complexity, background clutter, and frequent occlusions. The framework adopts an event-level formulation on a multi-task spatiotemporal architecture that combines a Vision Transformer (ViT) backbone for spatial feature extraction with a unidirectional Long Short-Term Memory (LSTM) network for temporal modeling over short video windows. The model jointly predicts handover occurrence and directionality, distinguishing between assistant-to-surgeon and surgeon-to-assistant transfers.

To improve robustness under visually complex surgical conditions, the framework incorporates data augmentation strategies as a preprocessing step during training to reduce the influence of irrelevant visual cues. Sequence-level predictions are aggregated into event-level detections, enabling clinically meaningful evaluation aligned with the perceptual characteristics of handover events. For comparative analysis, the proposed approach is evaluated against a VideoMamba-based temporal architecture designed for long-context sequence modeling.

Finally, we employ the Layer-CAM attribution method~\cite{jiang2021layercam} to analyze the spatial regions contributing to model decisions. This analysis provides insight into the visual cues driving predictions. Such interpretability is particularly valuable in surgical scenes, where critical regions, such as hands and instruments, often occupy small spatial extents and may be partially occluded.

Experimental evaluation on intraoperative videos from kidney transplant procedures demonstrates strong performance for both detection and direction classification tasks. The proposed approach achieves an event-level detection F1-score of 0.84 and a mean direction classification F1-score of 0.72. Compared with a VideoMamba-based baseline, the method achieves comparable detection performance while providing improved direction classification accuracy.
Our contributions are twofold:
\begin{itemize}
    \item \textbf{An event-level formulation for surgical instrument handover detection validated on intraoperative videos from real surgical procedures}, enabling robust analysis despite severe occlusions, background clutter, and temporal ambiguity.

    \item \textbf{A multi-task ViT--LSTM architecture} for jointly modeling handover occurrence and directionality, together with a systematic comparison against a VideoMamba-based temporal model to examine the impact of different temporal modeling strategies. We further provide an interpretable analysis of model behavior using the Layer-CAM attribution method, offering insight into the visual cues driving predictions and grounding the architectural findings in spatially meaningful evidence.
\end{itemize}

\section{Related Work}

Recent advances in artificial intelligence have significantly influenced medical imaging and surgical workflow analysis. Vision-based models have been widely explored for understanding clinical environments, including radiological interpretation, procedural analysis, and surgical activity recognition. Large-scale pretrained architectures and transformer-based models have demonstrated strong representation learning capabilities, motivating their adoption in complex medical visual domains.

Computer vision techniques have been extensively applied to surgical workflow understanding, including phase recognition, tool detection, and action recognition~\cite{twinanda2017endonet,Jin_SV_SCNET, nwoye2021}. EndoNet demonstrated early success in jointly modeling surgical phases and tool presence from laparoscopic videos~\cite{twinanda2017endonet}. Subsequent work emphasized higher-level interaction reasoning, including recognition of instrument–action–target relationships to capture fine-grained surgical activities~\cite{nwoye2021}. Prior studies have further explored surgical phase and action recognition using deep architectures and attention-based mechanisms~\cite{twinanda2017endonet, lea2016segmental,funke2019using}.

Beyond individual actions, group activity recognition in operating room environments has gained attention, where interactions among multiple clinical actors must be modeled under severe occlusions and visual clutter~\cite{operating_room_surveillance2023, Czempiel2021opera, hamoud2023, chen2025}. These works highlight the complexity of surgical scene understanding and the need for models capable of capturing both spatial structure and interaction dynamics.

Transformer-based architectures\cite{vaswani2017attention} have shown strong performance in visual representation learning by modeling long-range spatial dependencies through self-attention mechanisms~\cite{dosovitskiy2021vit, zhang2021swin}. ViTs have been successfully adopted in medical imaging tasks, improving contextual modeling and robustness in visually complex settings~\cite{hatamizadeh2022unetr,valanarasu2021medical, chen2021}. More recently, transformer-based architectures have been explored for surgical video understanding and workflow analysis. Approaches such as Trans-SVNet \cite{gao2021transsvnet} leverage hybrid spatial–temporal transformer embeddings for surgical phase recognition, while models like LoViT \cite{liu2023lovit} incorporate long-range temporal attention to capture dependencies across extended surgical procedures. ViT models have also been applied to decode intraoperative surgical activity directly from video recordings \cite{kiyasseh2023}.

Temporal reasoning is central to surgical video understanding, as clinically meaningful events unfold over time. Recurrent neural networks, particularly LSTM networks, have been widely used for modeling sequential dependencies in medical video data~\cite{hochreiter1997lstm, twinanda2017endonet, Jin_SV_SCNET}. Alternative approaches such as Temporal Convolutional Networks (TCNs) employ hierarchical temporal convolutions for action segmentation and detection~\cite{lea2017tcn}. Temporal modeling strategies have also been investigated for surgical phase segmentation and workflow analysis~\cite{zhang2023, czempiel2020tecno}. More recently, structured state space models have been proposed for efficient long-range sequence modeling~\cite{gu2022s4, gu2024mamba}. In the video domain, VideoMamba extends these ideas to enable efficient modeling of extended temporal contexts~\cite{li2024videomamba}. These methods provide complementary perspectives to recurrent approaches for capturing temporal dynamics.

Large pretrained vision–language models such as CLIP enable alignment between visual patterns and semantic concepts~\cite{clip2021, Zhao2023CLIPIM}. Prior work demonstrated the utility of CLIP-based representations in medical imaging and multimodal reasoning tasks~\cite{Zhao2023CLIPIM, wang-etal-2022-medclip}. The SurgiGuard framework \cite{SurgiGuard} combined CLIP-based visual representations with knowledge graph reasoning to detect surgical instrument handovers and model relationships among surgical entities. While this approach demonstrated the feasibility of automated interaction detection and visualization, its reliance on frame-level reasoning limited explicit modeling of temporal dynamics.

Over the years, numerous explanation methods across various categories have been proposed and extensively studied to unravel the complexities of black-box models \cite{Samek2021}. These include interpretable local surrogates \cite{localsurrogates}, occlusion analysis \cite{Zeiler_occlusion}, gradient-based methods (e.g., SmoothGrad \cite{smilkov2017_smoothgrad} and Integrated Gradients \cite{Sundararajan2017_Integrad}), and Layer-wise Relevance Propagation (LRP) \cite{Bach_lrp_2015}. However, user-centered evaluations demonstrate that while these methods may highlight different image regions, they can provide humans with comparable levels of understanding for image classification tasks \cite{dawoud2023}. The significance of such methods is particularly evident in clinical AI systems, where interpretability is essential for trustworthy decision-making \cite{SALVI2025}. For instance, gradient-based attribution techniques such as Grad-CAM localize image regions pertinent to model decisions~\cite{selvaraju2017gradcam}, while Layer-CAM refines spatial localization by utilizing activation maps across multiple layers \cite{jiang2021layercam}. Furthermore, recent work underscores the importance of interpretability in medical imaging systems, especially under conditions of uncertainty and occlusion. In particular, novel hybrid approaches that integrate explainability with uncertainty quantification have been proposed to enhance clinical credibility and decision robustness \cite{SALVI2025}. In this work, we apply Layer-CAM and a variation of the Integrated Gradients method to explain the spatio-temporal relevance of sequence inputs to the VideoMamba and our proposed model architecture, demonstrating that these models attend to critical regions of the input sequence that determine whether an instrument handover occurred and its direction.

Building upon prior vision-based approaches for surgical interaction analysis, the proposed framework introduces explicit spatiotemporal modeling for surgical instrument handover detection. In contrast to the SurgiGuard framework \cite{SurgiGuard}, which primarily relies on frame-level CLIP representations combined with graph-based reasoning, our approach integrates a ViT backbone with LSTM-based temporal modeling to jointly capture spatial features and temporal dependencies across video frames. Additionally, by adopting an event-level formulation and incorporating Layer-CAM attribution techniques, the framework facilitates interpretable analysis of handover events in intraoperative videos recorded during real surgical procedures.

\section{The Proposed Method}

\begin{figure*}[t]
  \centering
  \includegraphics[width=0.9\linewidth]{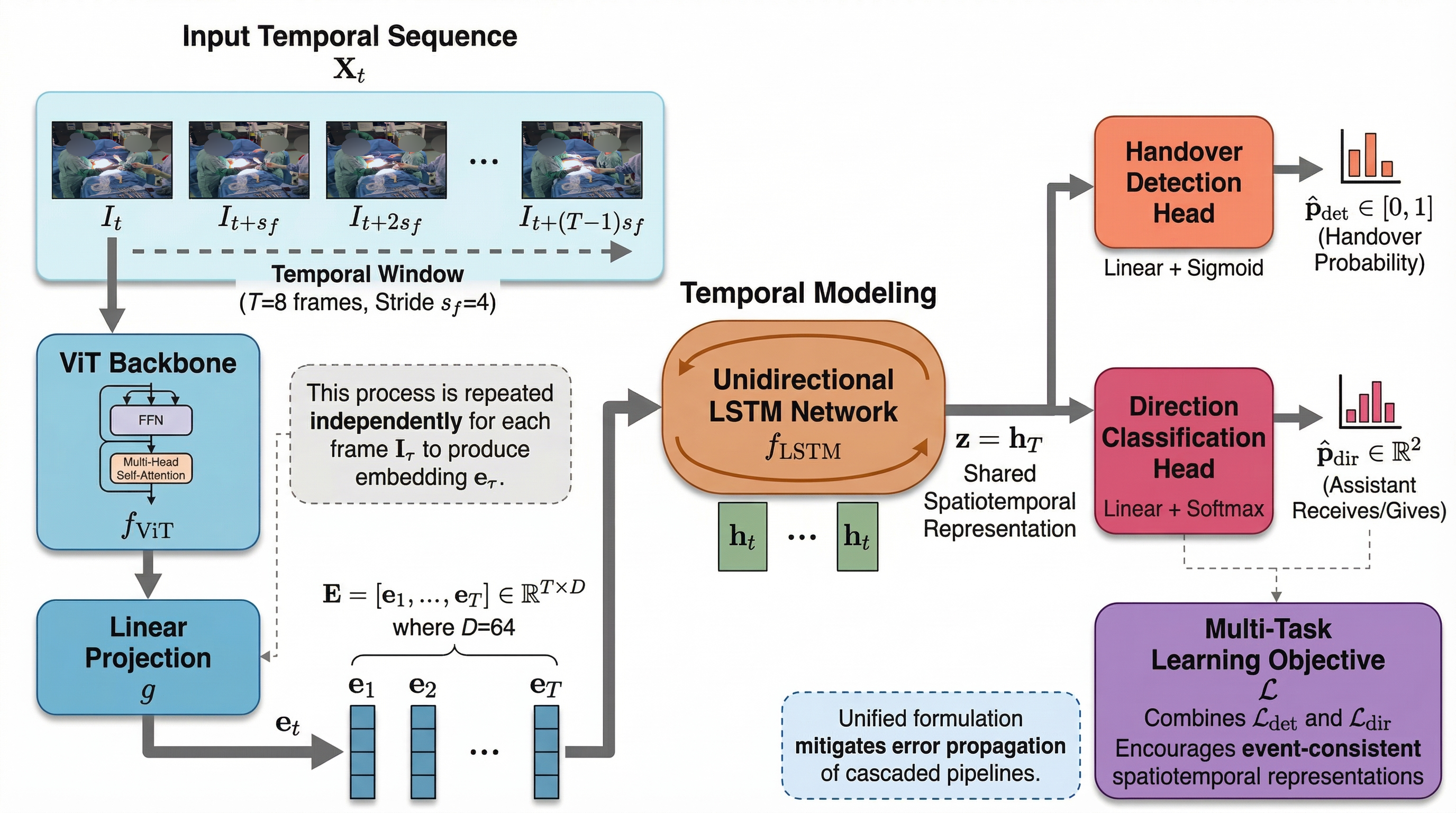}
\caption{
Overview of the proposed architecture. 
Spatial features are extracted independently from sampled video frames using a ViT backbone, 
projected into a compact embedding space, and temporally aggregated via a unidirectional LSTM. 
The shared spatiotemporal representation is jointly optimized using task-specific heads for handover detection 
and direction classification.
}
  \label{fig:vitals_architecture}
\end{figure*}

We address surgical instrument handover analysis as a joint detection and classification problem over short temporal windows. 
Rather than treating handover detection and direction estimation as separate stages, we adopt a unified multi-task formulation 
that jointly optimizes both objectives. As illustrated in Fig.~\ref{fig:vitals_architecture}, the proposed architecture combines transformer-based spatial 
feature extraction with recurrent temporal modeling. Given a sequence of sampled video frames, spatial representations are 
first computed independently using a ViT backbone. The resulting frame-level features are then projected 
into a compact embedding space and processed by a unidirectional LSTM network to capture temporal 
dependencies. The final LSTM representation is shared by two task-specific prediction heads corresponding to binary handover detection 
and direction classification. This design encourages the learning of event-consistent spatiotemporal representations while 
mitigating the error propagation effects commonly observed in cascaded pipelines.

\subsection{Problem Formulation}

Let a video be represented as a sequence of RGB frames 
$\mathcal{V} = \{\mathbf{I}_t\}_{t=1}^{N}$, where 
$\mathbf{I}_t \in \mathbb{R}^{H \times W \times 3}$ denotes the frame at time $t$.

For each time index $t$, we construct a temporal input sequence by sampling 
$T=8$ frames from the video with a fixed temporal stride $s_f=4$:

\begin{equation}
\mathbf{X}_t =
\{\mathbf{I}_t, \mathbf{I}_{t+s_f}, \mathbf{I}_{t+2s_f}, \dots, \mathbf{I}_{t+(T-1)s_f}\}.
\end{equation}

Each sequence therefore spans a temporal context of $(T-1)s_f + 1 = 29$ frames.
Each video frame is associated with a label $y \in \{0,1,2\}$ corresponding to the classes
\textit{assistant receives}, \textit{assistant gives}, and \textit{assistant idle}, where the latter refers
to frames in which the assistant is neither receiving nor giving an instrument.
A majority vote is taken over the five central frames of a sequence to determine its label.
This labeling scheme ensures that the model observes sufficient
context around the interaction, as well as the interaction itself.

\subsection{Spatial Feature Extraction}

Each sampled frame within a temporal window is independently encoded
using a ViT. Given an input frame $\mathbf{I}_\tau$,
the backbone first extracts a high-dimensional feature vector, which is
then projected into a compact embedding space for temporal modeling:

\begin{equation}
\mathbf{e}_\tau = g\!\left(f_{\text{ViT}}(\mathbf{I}_\tau)\right),
\quad \mathbf{e}_\tau \in \mathbb{R}^{D},
\end{equation}

where $f_{\text{ViT}}(\cdot)$ denotes the ViT backbone, $g(\cdot)$ is a
learnable linear projection layer, and $D=64$ is the embedding dimension.
Dropout regularization is applied to the projected embeddings to improve
generalization.  During training, the first 18 transformer layers of the ViT backbone are
frozen, while the higher layers are fine-tuned to adapt the spatial
representations to the handover analysis task.

For a temporal window of length $T$, the resulting embedding sequence is
\begin{equation}
\mathbf{E} =
[\mathbf{e}_1, \mathbf{e}_2, \dots, \mathbf{e}_T]
\in \mathbb{R}^{T \times D}.
\end{equation}

\subsection{Temporal Modeling}

Temporal dependencies are modeled using a unidirectional LSTM network. This design choice reflects the statistical characteristics of intraoperative video data, where annotated datasets 
are typically limited in scale and exhibit sparse event distributions. While transformer-based temporal models 
offer high representational capacity, they rely on weak sequential inductive bias and often require substantially 
larger datasets to ensure stable optimization.
In contrast, recurrent architectures impose an explicit temporal ordering constraint and provide a strong sequential 
inductive bias, which is well-suited for modeling short interaction sequences with limited temporal diversity. 
Moreover, the proposed framework delegates complex spatial dependency modeling to the ViT backbone, 
allowing the LSTM to focus on lightweight temporal aggregation of frame-level embeddings. This separation reduces 
model complexity while preserving sensitivity to the temporal evolution of handover events.
At each time step $t$, the LSTM updates its hidden state:

\begin{equation}
\mathbf{h}_t = f_{\text{LSTM}}(\mathbf{e}_t, \mathbf{h}_{t-1}),
\end{equation}

where $\mathbf{h}_t \in \mathbb{R}^{H}$ denotes the hidden state. The final hidden state summarizes the temporal evolution of the sampled window 
and serves as a compact sequence-level representation:

\begin{equation}
\mathbf{z} = \mathbf{h}_T.
\end{equation}


\subsection{Multi-Task Prediction Heads}

The shared representation $\mathbf{z}$ is passed to two task-specific prediction heads.

\paragraph{Handover Detection Head}

We model handover detection as a binary classification problem:

\begin{equation}
\hat{p}_{\text{det}} = \sigma(\mathbf{W}_{\text{det}} \mathbf{z} + \mathbf{b}_{\text{det}}),
\end{equation}

where $\sigma(\cdot)$ denotes the sigmoid function.

\paragraph{Direction Classification Head}

Direction classification is formulated as a categorical prediction:

\begin{equation}
\hat{\mathbf{p}}_{\text{dir}} = 
\text{softmax}(\mathbf{W}_{\text{dir}} \mathbf{z} + \mathbf{b}_{\text{dir}}),
\end{equation}

where $\hat{\mathbf{p}}_{\text{dir}} \in \mathbb{R}^{2}$ corresponds to 
\textit{assistant receives} and \textit{assistant gives}.

\subsection{Learning Objective}

The network is trained via a multi-task objective combining detection and direction losses.

\paragraph{Detection Loss}

The binary detection label $y_{\text{det}}$ is derived from the original label $y$ as

\begin{equation}
    y_{\text{det}} = \mathbbm{1}\left[y \in \{0,1\}\right],
\end{equation}

where $\mathbbm{1}[\cdot]$ denotes the indicator function.
To address class imbalance between handover and non-handover windows, 
we employ weighted binary cross-entropy:

\begin{equation}
\mathcal{L}_{\text{det}} =
- w_{pos} y_{\text{det}} \log \hat{p}_{\text{det}}
- (1 - y_{\text{det}}) \log (1 - \hat{p}_{\text{det}}),
\end{equation}

where $w_{pos}$ controls the contribution of  positive samples.

\paragraph{Direction Loss}

Direction supervision is applied only to positive detection samples, i.e., when $y_{\text{det}}=1$.
For such samples, the direction loss is

\begin{equation}
\mathcal{L}_{\text{dir}}
=
\mathrm{WCE}\!\left(\hat{\mathbf{p}}_{\text{dir}}, y_{\text{dir}}\right),
\end{equation}

where $\mathrm{WCE}(\cdot)$ denotes the weighted cross-entropy loss.
The direction label is derived from the ground-truth class label as
\(
y_{\text{dir}} = y.
\)

\paragraph{Total Loss}

The full training objective is:

\begin{equation}
\mathcal{L} =
\lambda_{\text{det}} \mathcal{L}_{\text{det}} +
\lambda_{\text{dir}} \mathcal{L}_{\text{dir}},
\end{equation}

where $\lambda_{\text{det}}$ and $\lambda_{\text{dir}}$ balance the contributions
of detection and direction objectives. The multi-task formulation encourages the learning of shared spatiotemporal
representations that capture both event occurrence and interaction semantics,
while reducing error propagation effects typical of cascaded pipelines.

\subsection{VideoMamba Comparison Model}

To contextualize the proposed Multi-Task ViT--LSTM architecture, we evaluated a comparison model based on VideoMamba~\cite{li2024videomamba}, a state-space model designed for efficient temporal modeling in video. In contrast to transformer-based temporal attention, whose cost grows quadratically with sequence length, VideoMamba relies on structured state-space operators with approximately linear complexity, enabling a longer temporal context at comparable compute.

To be comparable to the Multi-Task ViT--LSTM, we construct input clips of 8 frames sampled with a temporal stride of 4 (covering 29 frames of temporal extent). Two VideoMamba models were trained separately: One instance was optimized for the prediction of general handover events, while the second model was specifically trained to infer the directional outcome within a given handover sequence.

We employ a VideoMamba-Middle backbone pretrained on Kinetics-400\cite{kay2017kineticshumanactionvideo}. The original classification head is replaced with a lightweight projection head (four linear layers) mapping the 576-dimensional CLS token to one output prediction. During fine-tuning, only the last 12 of 24 Mamba blocks are updated while earlier blocks remain frozen. Clip-level targets are assigned using the center frame label; during training, we apply a relaxed majority-vote labeling within a $\pm 2$ frame neighborhood to reduce sensitivity to annotation jitter near event boundaries.

\section{Experiments}

\paragraph{Preprocessing.}

To reduce label noise and improve annotation reliability, candidate handover instances were excluded from training if they exhibited: (1) ambiguous interactions,  
(2) severe occlusions preventing reliable interpretation, or  
(3) multiple simultaneous handovers within the same temporal interval.
\paragraph{Dataset Details.}

Dataset construction followed a two-stage procedure. First, temporal segments 
containing surgical instrument handover interactions were manually extracted 
from continuous intraoperative recordings. Second, the extracted segments were 
annotated at the frame level with labels specifying both handover occurrence 
and directionality. 
The final dataset consists of 484 annotated handover events in both directions, 
including 334 assistant-to-surgeon and 150 surgeon-to-assistant interactions, collected from five real kidney transplant surgeries. 
One surgery and its corresponding 50 handover events are reserved as a test set, while the remaining 434 events are used for model training and validation. 

\paragraph{Evaluation Protocol.}
We evaluate the models along two complementary dimensions:
(i) handover detection and (ii) handover direction classification.

\begin{figure}[tp]
    \centering
    \includegraphics[width=0.9\linewidth]{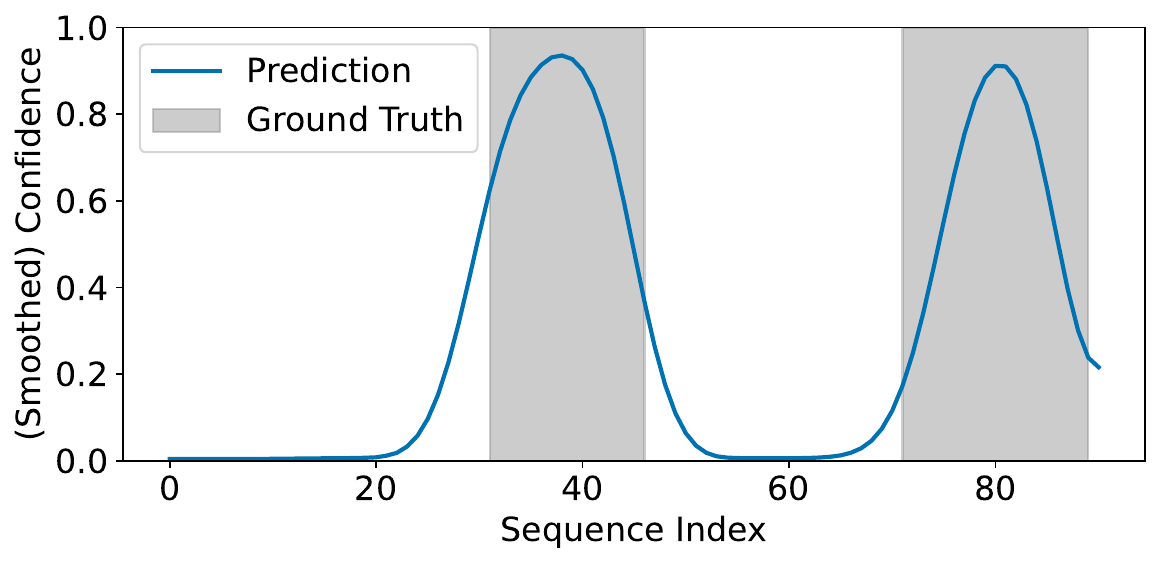}
    \caption{Handover detection evaluation approach applied to a video snippet from an actual surgical procedure, which includes two handover events. The smoothed temporal signal is depicted in blue and the grey shaded region indicates the ground truth, defined as the sequences containing at least one handover frame.}
    \label{fig:prediction-sequence-example}
\end{figure}

Detection is treated as a temporal event localization problem operating on sequence-level confidence scores, while direction is evaluated independently of detection to avoid compounding errors. For detection, model outputs are converted into a one-dimensional temporal confidence signal over time (Fig.~\ref{fig:prediction-sequence-example}). 
Each prediction corresponds to an input sequence of eight frames sampled with a frame stride of four, while consecutive sequences are generated with a sequence stride of two. 
For the evaluation, a sequence is labeled as a handover if it contains at least one handover frame. The temporal confidence signal is smoothed using a Gaussian filter ($\sigma=3$, kernel size $=15$) to suppress noise and stabilize local maxima. 
To reduce boundary artifacts during filtering, reflective padding of $3\sigma$ is applied. 
Handover candidates are then extracted using prominence-based peak detection with a minimum peak height of $0.1$ and a prominence threshold of 1\% of the signal range for each extracted video segment. Detected peaks are matched to ground-truth handover events using a temporal tolerance of two sequences around the annotated handover interval. 
This tolerance accounts for minor temporal misalignment introduced by sequence sampling and temporal smoothing.

For direction evaluation, predictions are aggregated within each ground-truth handover interval using Gaussian-weighted temporal aggregation. 
This weighting emphasizes predictions near the center of the event while reducing the influence of temporally misaligned or boundary-near predictions. 
The aggregated confidence produces a single direction prediction per handover event, which is thresholded at 0.5 to determine the final handover direction.

\begin{table}[t]
\centering
\caption{Classification metrics for handover detection and direction performance (50 handovers). 
Detection uses Gaussian smoothing and prominence-based peak detection with a temporal tolerance of two windows. 
Direction is evaluated independently using Gaussian-weighted aggregation within ground-truth intervals. F1@R and F1@G denote the class-wise F1 scores for the assistant-receives and assistant-gives directions, respectively.}
\label{tab:combined_results}
\begin{tabular}{lccc}
    \hline
    \textbf{Detection Performance} \\
    \hline
    Model & Precision & Recall & F1 \\
    \hline
    Multi-task ViT--LSTM & 0.81 & 0.86 & 0.84 \\
    Single-task ViT--LSTM & 0.97 & 0.66 & 0.79 \\
  
    VideoMamba     & 0.82 & 0.86 & 0.84 \\ 
    \hline
    \textbf{Direction Performance} \\
    \hline
    Model & F1@R & F1@G & Mean \\
    \hline
    Multi-task ViT--LSTM & 0.70 & 0.74 & 0.72 \\
    Single-task ViT--LSTM & 0.55 & 0.70 & 0.63 \\
    VideoMamba & 0.60 & 0.61 & 0.61 \\
   
    \hline
\end{tabular}
\end{table}

\paragraph{Detection Performance.}

Detection results are summarized in Table \ref{tab:combined_results}.
All evaluated models operate on the same temporal input consisting of eight sampled frames per sequence, enabling a consistent comparison across architectures. The multi-task ViT–LSTM achieves a detection F1 score of 0.84, with a balanced precision–recall trade-off (precision 0.81, recall 0.86). The single-task ViT–LSTM achieves a higher precision of 0.97 but a substantially lower recall of 0.66, resulting in an overall F1 score of 0.79. This behavior indicates that optimizing solely for detection encourages more conservative predictions that reduce false positives but miss a larger number of true handover events.

The VideoMamba model achieves a detection F1 score of 0.84, matching the multi-task ViT–LSTM. Its performance is characterized by a balanced precision–recall profile (0.82 precision, 0.86 recall), suggesting that the state-space temporal modeling employed by VideoMamba effectively captures the temporal dynamics of surgical instrument exchanges while maintaining stable event localization.

\paragraph{Direction Performance.}

\begin{figure}[t]
\centering
\includegraphics[width=\linewidth]{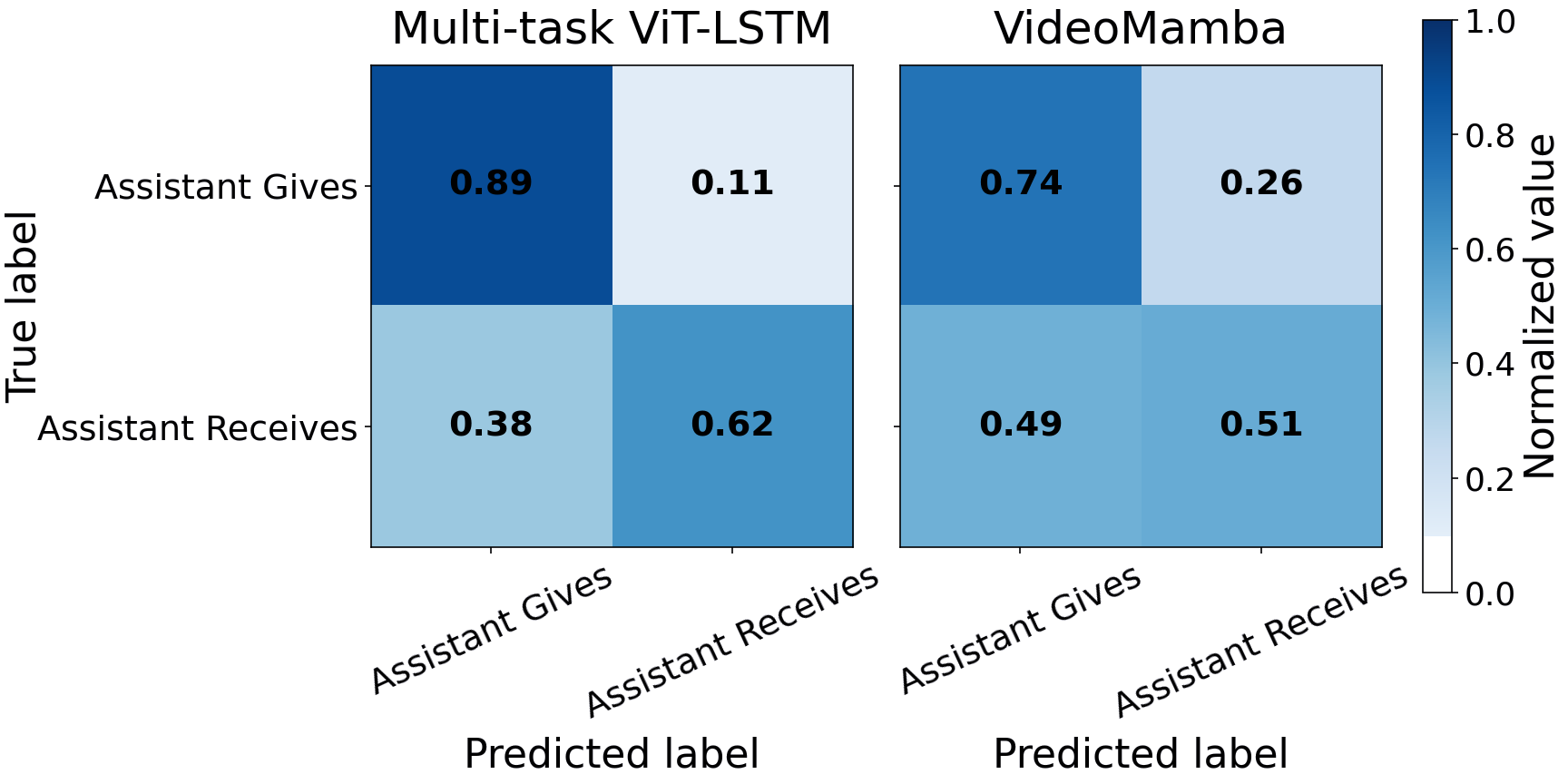}
\caption{Normalized confusion matrices comparing multi-task ViT--LSTM and VideoMamba.}
\label{fig:direction_confusion}
\end{figure}

Direction classification results are summarized in Table~\ref{tab:combined_results}, with normalized confusion matrices for the multi-task ViT--LSTM and VideoMamba shown in Fig.~\ref{fig:direction_confusion}. Among the evaluated models, the multi-task ViT--LSTM achieves the best overall performance, with a mean F1 score of 0.72, including 0.70 for assistant-receives interactions and 0.74 for assistant-gives interactions. Table~\ref{tab:combined_results} shows that performance is relatively balanced across both interaction types, although assistant-receives events remain slightly more difficult to classify correctly.

For the single-task ViT--LSTM, the direction model is obtained by sequential fine-tuning from the detection checkpoint. This model achieves F1@R = 0.55 and F1@G = 0.70, with a mean F1 score of 0.63, where F1@R and F1@G denote the class-wise F1 scores for the assistant-receives and assistant-gives directions, respectively. The VideoMamba direction model is also initialized from the detection checkpoint and further fine-tuned for direction classification. It achieves F1@R = 0.60 and F1@G = 0.61, with a mean F1 score of 0.61, showing more balanced performance between the two interaction types than the single-task ViT--LSTM, although its overall F1 score remains lower than that of the multi-task model.

Overall, the normalized confusion matrices for the multi-task ViT--LSTM and VideoMamba shown in Fig. \ref{fig:direction_confusion} confirm that direction classification remains challenging, while also highlighting the stronger recall achieved by the multi-task model. Since the matrices are normalized with respect to the ground-truth labels, the diagonal entries correspond to class-wise recall. The multi-task ViT--LSTM achieves recall values of 62\% for assistant-receives and 89\% for assistant-gives, whereas VideoMamba achieves 51\% and 74\%, respectively. Thus, assistant-gives is recognized more reliably than assistant-receives in both models, and the multi-task formulation improves recall for both interaction directions.  This lower performance likely reflects both the smaller number of training instances available for direction classification and the visual similarity of hand trajectories during instrument transfer, particularly under partial occlusions, motion blur, and limited visibility of the exchanged instrument in real surgical scenes.

\begin{figure}[b]
    \centering
    \begin{subfigure}{0.48\linewidth}
        \centering
        \includegraphics[width=\linewidth]{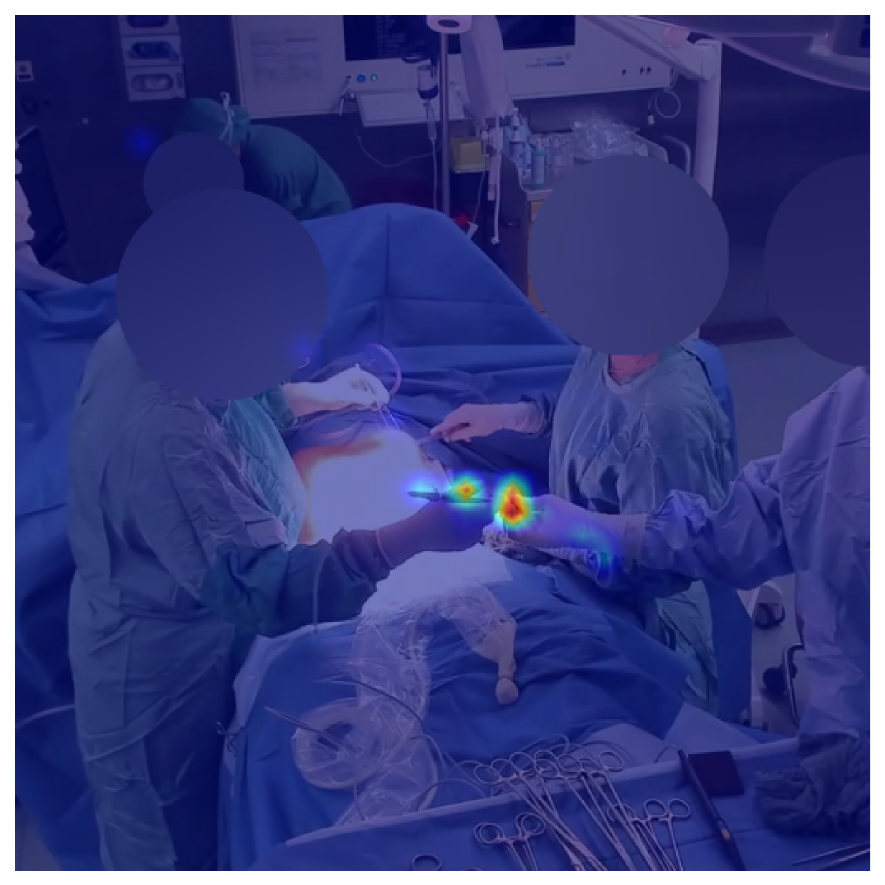}
        \caption{Training Set}
        \label{fig:layer-cam-single-frame-train}
    \end{subfigure}
    \hfill
    \begin{subfigure}{0.48\linewidth}
        \centering
        \includegraphics[width=\linewidth]{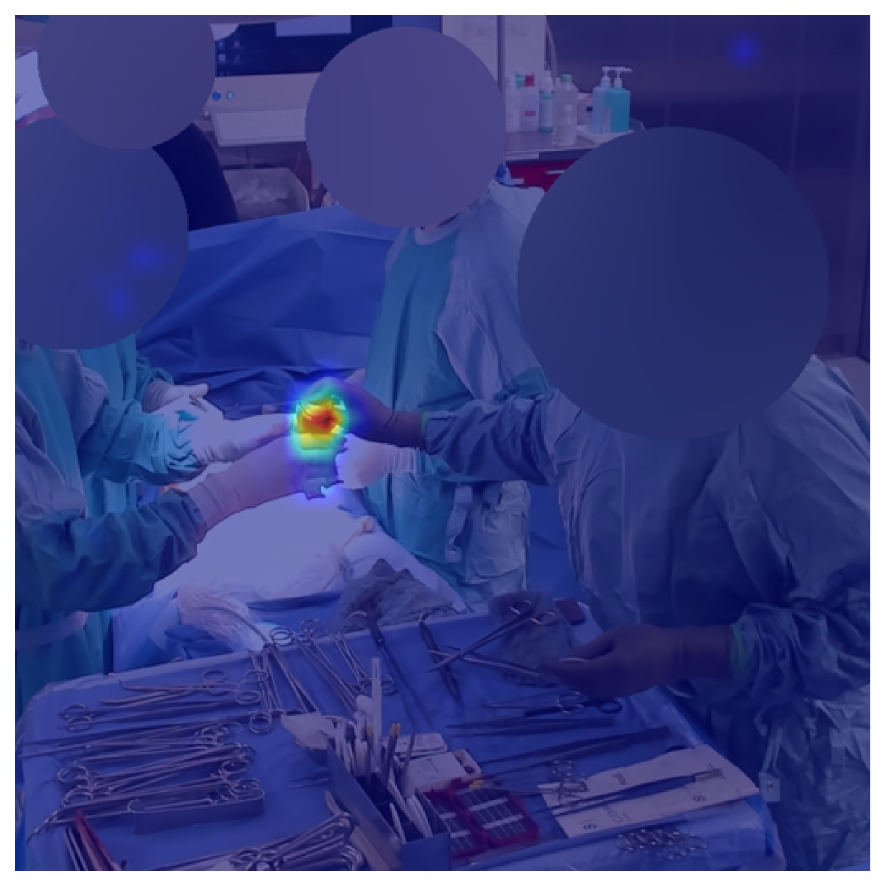}
        \caption{Test Set}
        \label{fig:layer-cam-single-frame-test}
    \end{subfigure}
    \caption{Layer-CAM explanation maps of the multi-task ViT--LSTM model illustrating the contribution of individual frames to the handover detection prediction. Frame \ref{fig:layer-cam-single-frame-train} is extracted from a sequence belonging to the training set, whereas frame \ref{fig:layer-cam-single-frame-test} is taken from a sequence in the test set. 
    }\label{fig:layer-cam-single-frame}
\end{figure}

\begin{figure}[t]
    \centering
    \begin{subfigure}{0.48\linewidth}
        \centering
        \includegraphics[width=\linewidth]{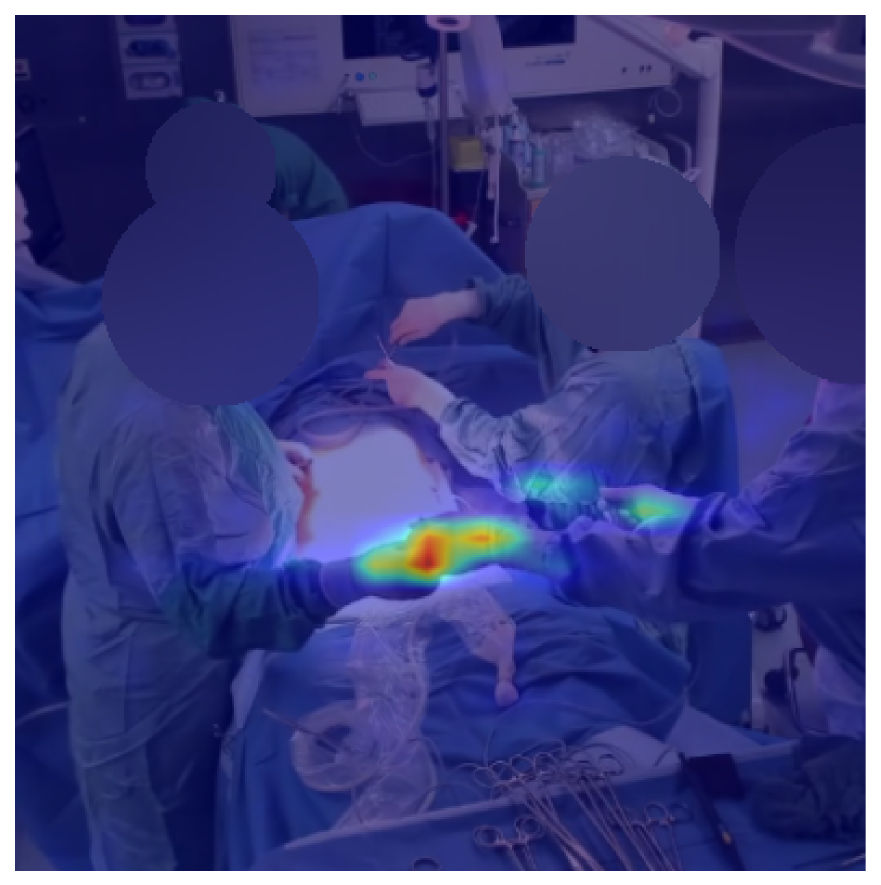}
        \caption{Training Set}
        \label{fig:vm-gradient-viz-train}
    \end{subfigure}
    \hfill
    \begin{subfigure}{0.48\linewidth}
        \centering
        \includegraphics[width=\linewidth]{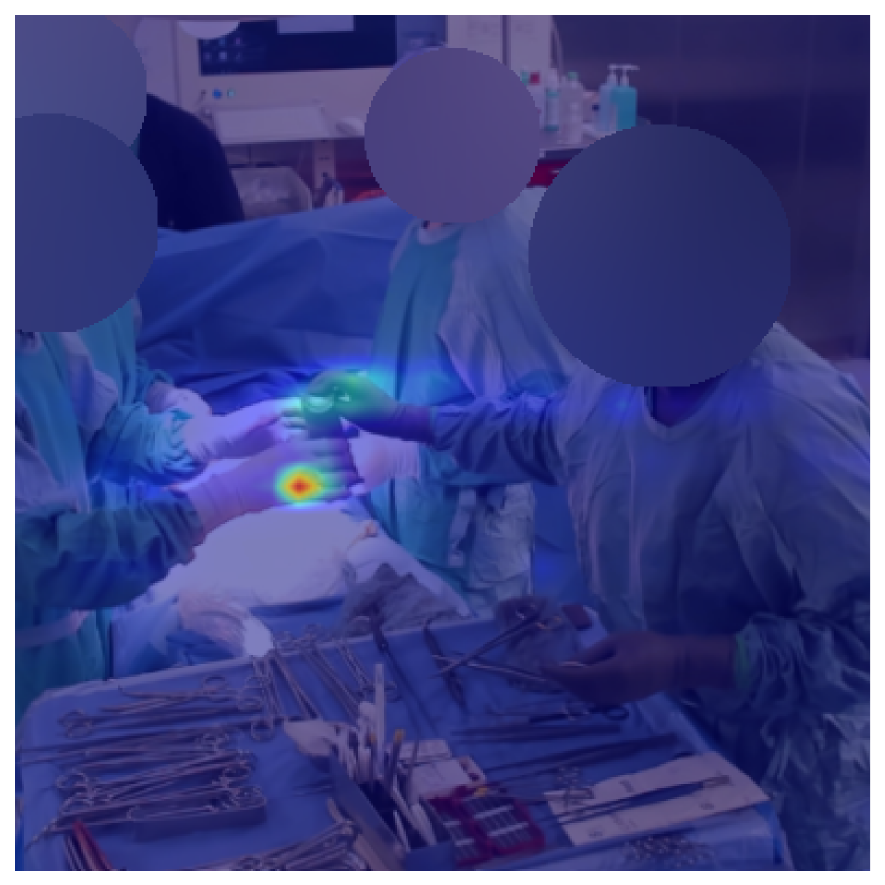}
        \caption{Test Set}
        \label{fig:vm-gradient-viz-val}
    \end{subfigure}
    \caption{Visualized gradient accumulation of the VideoMamba model demonstrates its ability to recognize the significance of hands during the handover process. The gradient accumulations were collected on a per-pixel basis and subsequently summed by patch to reduce noise.}
    \label{fig:vm-gradient-viz}
\end{figure}

\paragraph{Computational Complexity.}
To complement the performance evaluation, Table~\ref{tab:model_efficiency} reports the computational characteristics of the compared architectures using the same 8-frame input sequence.  The multi-task ViT--LSTM model contains approximately 304M parameters, primarily due to the large ViT backbone used for spatial feature extraction. This architecture also incurs substantially higher computational cost and inference latency. In particular, the self-attention mechanism in ViTs scales quadratically with the number of visual tokens (image patches), leading to very high FLOPs when processing high-resolution inputs. In contrast, the VideoMamba model requires considerably fewer parameters (74M) and exhibits significantly lower computational cost and latency. Its state-space temporal modeling enables efficient sequence processing while maintaining linear complexity with respect to sequence length. Overall, these results highlight the trade-off between the strong spatial representation capacity of the ViT-based architecture and the computational efficiency of the VideoMamba model.
\paragraph{Interpretability Analysis.}
To investigate whether the proposed model relies on semantically meaningful visual cues, we analyzed spatial attribution maps generated using Layer-CAM across both training and held-out test sequences. As illustrated in Fig.~\ref{fig:layer-cam-single-frame}, the explanation maps reveal that the model consistently attends to hand–instrument interaction regions throughout the temporal window, with activation patterns concentrated around the exchange zone where the handover occurs. Notably, this localized attention behavior is not restricted to training samples but is also observed on the unseen test set from a different surgery. 
This consistency suggests that the learned representations capture transferable visual patterns associated with handover dynamics rather than overfitting to surgery-specific appearance characteristics. Although the current size of the dataset constrains the extent of generalization analysis, these findings indicate that the model is learning in a clinically meaningful direction and would likely benefit from continued data acquisition over time. Similarly, gradient-based attribution applied to the VideoMamba model (shown in Fig.~\ref{fig:vm-gradient-viz}) demonstrates that it also attends to the relevant portions of the handover sequence, with accumulated gradients highlighting the frames and spatial regions where the instrument transfer unfolds. Together, these analyzes provide evidence that both architectures ground their predictions in visually interpretable interaction cues, supporting the validity of the learned representations for surgical handover analysis.

\begin{table}[t]
\centering
\footnotesize
\caption{Computational characteristics of the compared models with an input sequence length of 8. ViT--LSTM (image size of $518 \times 518$, and VideoMamba-Middle ($512 \times 512$). The total encoding time for eight images processed by the Vision Transformer (ViT) is 232 milliseconds. However, for real-time applications, the design can be improved by implementing sequential processing and storing the ViT encodings in a queue. As a result, processing each new image would take approximately 30 milliseconds, which is appropriate for real-time performance.}
\label{tab:model_efficiency}
\begin{tabular}{lccc}
\hline
Model &  Params & FLOPs & Latency (ms) \\ 
\hline
Multi-task ViT--LSTM &  304.47M & $3.32\times10^{12}$ & 232.61\\ 
VideoMamba-Middle & 74.32M & $7.25\times10^{9}$ & 61.12 \\
\hline
\end{tabular}
\end{table}

\section{Conclusion and Future Work}

We introduced an interpretable spatiotemporal vision framework for event-level analysis of surgical instrument handovers 
in intraoperative videos acquired during real surgical procedures. The proposed architecture combines transformer-based spatial 
representation learning with recurrent temporal aggregation and employs a unified multi-task formulation to jointly model handover 
occurrence and interaction directionality.

Evaluation on real surgical recordings demonstrates that explicit task separation and recurrent temporal modeling enable reliable handover detection despite severe occlusions, background clutter, and restricted 
visibility. The proposed approach achieves an event-level detection F1-score of 0.84 and a mean direction classification F1-score of 0.72 on kidney transplant procedures. Comparative analysis with a VideoMamba-based temporal architecture revealed complementary behaviors: while both models achieve comparable detection performance (F1 = 0.84), VideoMamba exhibits lower direction classification performance (mean F1 = 0.61). These observations underscore the importance of aligning temporal modeling strategies with both dataset characteristics and task structure.

Layer-CAM-based attribution analyses provided further insight into the visual mechanisms underlying model predictions. The resulting 
explanation maps indicate that decisions are primarily driven by localized hand–instrument interaction patterns. Such interpretability is particularly valuable in surgical 
applications, where understanding model behavior is essential for trust, validation, and potential clinical integration.

Several directions for future work emerge from this study. 
First, improving robustness under severe occlusions and viewpoint variability remains a major challenge in surgical video analysis. Future approaches may benefit from incorporating motion-aware representations, cross-frame correspondence modeling, or multi-view learning strategies that help mitigate partial observability in complex operating room environments. Second, expanding dataset scale and diversity across procedures, institutions, and surgical settings is essential for improving model generalization and reducing dataset-specific bias. As larger annotated intraoperative video collections become available, more expressive temporal modeling strategies could be explored. In particular, future work may further investigate the role of temporal modeling strategies for surgical interaction analysis. While the proposed multi-task ViT--LSTM framework demonstrates strong performance for both detection and direction classification, the comparison with the VideoMamba-based architecture highlights how different temporal modeling paradigms capture complementary aspects of the interaction dynamics. The use of an LSTM for temporal aggregation in the present work reflects the limited size and sparsity of currently available annotated surgical video datasets, where recurrent models provide a useful sequential inductive bias and stable optimization behavior. As larger and more diverse surgical video datasets become available, longer-context temporal architectures such as state-space models or transformer-based sequence models may become increasingly effective, enabling richer long-range interaction modeling and more flexible temporal context integration for surgical workflow analysis.

\newpage
{
    \small
    \bibliographystyle{ieeenat_fullname}
    \bibliography{manuscript_ref}
}


\maketitlesupplementary
\appendix
\noindent
This document provides additional material supporting the main paper. It includes a detailed description of the training procedure, implementation details of the evaluated models, and additional qualitative results illustrating model behavior. These supplementary analyses complement the experimental results presented in the main manuscript.

\section{Training Procedure}

Algorithm~\ref{alg:vitals} summarizes the training procedure of the proposed  framework.

\begin{algorithm}[h]
\caption{Training on windowed surgical video}
\label{alg:vitals}

\KwIn{Video frames $\{I_t\}_{t=1}^N$; sampled frames $T{=}8$; frame stride $s_f{=}4$; sequence stride $s_s{=}2$; labels $y \in \{0,1,2\}$}
\KwOut{Trained parameters of ViT backbone $f_{\mathrm{ViT}}$, projection $g$, LSTM $f_{\mathrm{LSTM}}$, detection head, direction head}

\ForEach{training iteration}{

    Sample sequence start indices $\{t_i\}_{i=1}^B$\;

    \tcp{Construct temporal input sequences}
    $X \leftarrow \{\,I_{t_i + k s_f}\,\}_{i=1,k=0}^{B,T-1}$ 
    \tcp*{$X \in \mathbb{R}^{B \times T \times H \times W \times 3}$}

    Apply training augmentation to all frames in $X$\;

    \tcp{Spatial encoding (per frame)}
    $F \leftarrow f_{\mathrm{ViT}}(X)$
    \tcp*{frame-wise ViT features}

    \tcp{Projection to temporal embedding space}
    $E \leftarrow g(F)$ 
    \tcp*{$E \in \mathbb{R}^{B \times T \times D}$}

    \tcp{Temporal aggregation}
    $z \leftarrow f_{\mathrm{LSTM}}(E)$ 
    \tcp*{final hidden state $z \in \mathbb{R}^{B \times H}$}

    \tcp{Predictions}
    $\hat{p}_{\mathrm{det}} \leftarrow \sigma(\mathrm{Head}_{\mathrm{det}}(z))$\;
    $\hat{\mathbf{p}}_{\mathrm{dir}} \leftarrow \mathrm{softmax}(\mathrm{Head}_{\mathrm{dir}}(z))$\;

    \tcp{Targets}
    $y_{\mathrm{det}} \leftarrow \mathbbm{1}[y \neq 2]$\;
    $m \leftarrow y_{\mathrm{det}}$\;
    $y_{\mathrm{dir}} \leftarrow y$\;

    \tcp{Multi-task loss}
    $\mathcal{L}_{\mathrm{det}} \leftarrow \mathrm{WBCE}(\hat{p}_{\mathrm{det}}, y_{\mathrm{det}})$\;
    $\mathcal{L}_{\mathrm{dir}} \leftarrow \mathrm{WCE}(\hat{\mathbf{p}}_{\mathrm{dir}}[m], y_{\mathrm{dir}}[m])$\;

    $\mathcal{L} \leftarrow \lambda_{\mathrm{det}}\mathcal{L}_{\mathrm{det}} + \lambda_{\mathrm{dir}}\mathcal{L}_{\mathrm{dir}}$\;

    Update parameters via backpropagation on $\mathcal{L}$\;
}
\end{algorithm}

\section{Implementation Details}
This section provides details on the models used in this work. 

\subsection{Multi-Task ViT-LSTM}
The Multi-Task ViT-LSTM uses a Pytorch Image Models checkpoint of a ViT-Large/14 backbone pretrained with the DINOv2 self-supervised method \cite{dinov2} on the LVD-142M dataset. Table \ref{tab:multi_task_vit_lstm_details} contains the detailed model architecture and training setup.

\begin{table}[t]
\centering
\small
\caption{Implementation details of ViT-LSTM model.}
\label{tab:multi_task_vit_lstm_details}
\begin{tabularx}{\columnwidth}{|>{\raggedright\arraybackslash}X|>{\raggedright\arraybackslash}X|}
    \hline
    \textbf{Quantity} & \textbf{Value} \\
    \hline
    Image Input Size & 518 x 518 \\
    \hline
    Feature Projection Dimension & 64 \\ 
    \hline
    Backbone Learning Rate & $3 \times 10^{-6}$ \\
    \hline
    Backbone Weight Decay &  $1 \times 10^{-4}$ \\
    \hline
    Backbone Layers Frozen & 18 of 24 \\
    \hline
    Backbone Ouput Dropout Rate     &  0.3 \\
    \hline
    LSTM Learning Rate & $1 \times 10^{-5}$ \\
    \hline
    LSTM Weight Decay &  $1 \times 10^{-5}$ \\
    \hline
    LSTM Hidden Size &  64 \\
    \hline
    LSTM Hidden Layers &  1 \\
    \hline
    LSTM Output Dropout Rate  &  0.4 \\
    \hline
    Batch Size & 8 \\
    \hline
    LR Scheduler & 5\% Linear Warmup \\
                & + Cosine Annealing \\
    \hline
    Gradient Accumulation Steps & 2 \\
    \hline
    Effective Batch Size & 16 \\
    \hline
    Max Gradient Norm & 1.0 \\
    \hline
    Loss Weighting & $w_{pos}=1.5$ \\
                   & $\lambda_{\text{det}}=2.5$ \\
                   & $\lambda_{\text{dir}}=1$ \\
    \hline
    Number of Epochs (Incl. Early Stopping)    & 10 \\
    \hline
    Training Augmentations    & JitteredCenterCrop \\
                              & ColorJitter \\
                              & HorizontalFlip \\
    \hline
    Test Augmentations    & JitteredCenterCrop \\

    \hline
\end{tabularx}
\end{table}

The models are trained on a single NVIDIA RTX 6000 Ada GPU. Each epoch processes one third of the dataset. A weighted random sampler is used with fixed class sampling probabilities of 0.6, 0.2, and 0.2 for \textit{assistant idle}, \textit{assistant receives}, and \textit{assistant gives}, respectively. The AdamW optimizer is used. Regarding the tranformations, \textit{JitteredCenterCrop}  crops a fixed-size fraction of the image around the centre, but randomly jitters the crop centre within a specified horizontal and vertical range. More specifically, it crops 40\% of the image width and 71.1\% of the height around the centre, randomly shifting the crop by up to $\pm$3\% horizontally and $\pm$5\% vertically before ensuring the crop remains within the image boundaries. For testing, no shifting is applied. \textit{ColorJitter} is applied with brightness $\pm$0.2, contrast $\pm$0.1, saturation $\pm$0.2, and hue $\pm$0.05. The probability of \textit{HorizontalFlip} is 0.5.

\subsection{VideoMamba}
As a comparison model, we employ a VideoMamba backbone~\cite{li2024videomamba} pretrained on Kinetics-400 at $224 \times 224$ resolution with 8 input frames. The backbone's classification head is replaced by a custom projection head consisting of four linear layers with LayerNorm, GELU activations, and dropout (rate 0.3), mapping from the 576-dimensional CLS token to the three handover classes.

\begin{table}[t]
\centering
\small
\caption{Implementation details of the VideoMamba comparison model.}
\label{tab:videomamba_details}
\begin{tabularx}{\columnwidth}{|>{\raggedright\arraybackslash}X|>{\raggedright\arraybackslash}X|}
    \hline
    \textbf{Quantity} & \textbf{Value} \\
    \hline
    Backbone Variant & VideoMamba-Middle \\
    \hline
    Pretraining Dataset & Kinetics-400 \\
    \hline
    Image Input Size & 512 $\times$ 512 \\
    \hline
    Number of Input Frames & 8 \\
    \hline
    Backbone Embedding Dimension & 576 \\
    \hline
    Backbone Layers Frozen & 12 of 24 \\
    \hline
    Backbone Learning Rate & $5 \times 10^{-5}$ \\
    \hline
    Projection Head Learning Rate & $1 \times 10^{-4}$ \\
    \hline
    Weight Decay & $5 \times 10^{-4}$ \\
    \hline
    Projection Head Architecture & 576--256--128--128--1 \\
    \hline
    Projection Head Dropout Rate & 0.3 \\
    \hline
    Drop Path Rate (Backbone) & 0.1 \\
    \hline
    LR Scheduler & CosineAnnealingWarmRestarts ($T_0{=}10$) \\
    \hline
    Training Label Strategy & Majority vote (5-frame window) \\
    \hline
    Loss Function & BCEWithLogits (weighted) \\
    \hline
    Training Augmentations    & JitteredCenterCrop \newline ColorJitter \newline HorizontalFlip \\
    \hline
    Test Augmentations    & JitteredCenterCrop \\
    \hline
\end{tabularx}
\end{table}

 We selectively fine-tune the last 12 of 24 Mamba blocks along with the final normalization layer, keeping the remaining backbone frozen. The backbone is trained with a learning rate of $5 \times 10^{-5}$ and the projection head at $1 \times 10^{-4}$, both using AdamW with a weight decay of $5 \times 10^{-4}$ and a cosine annealing schedule with warm restarts every 10 epochs. Classification is performed for the center frame of each clip via a relaxed majority-vote labeling over a 5-frame window during training.  More details are provided in Table~\ref{tab:videomamba_details}.


\section{Additional Figures} \label{app:additional-figures}
\begin{figure}[t!]
    \centering
    \begin{subfigure}{0.95\linewidth}
        \centering
        \includegraphics[width=\linewidth]{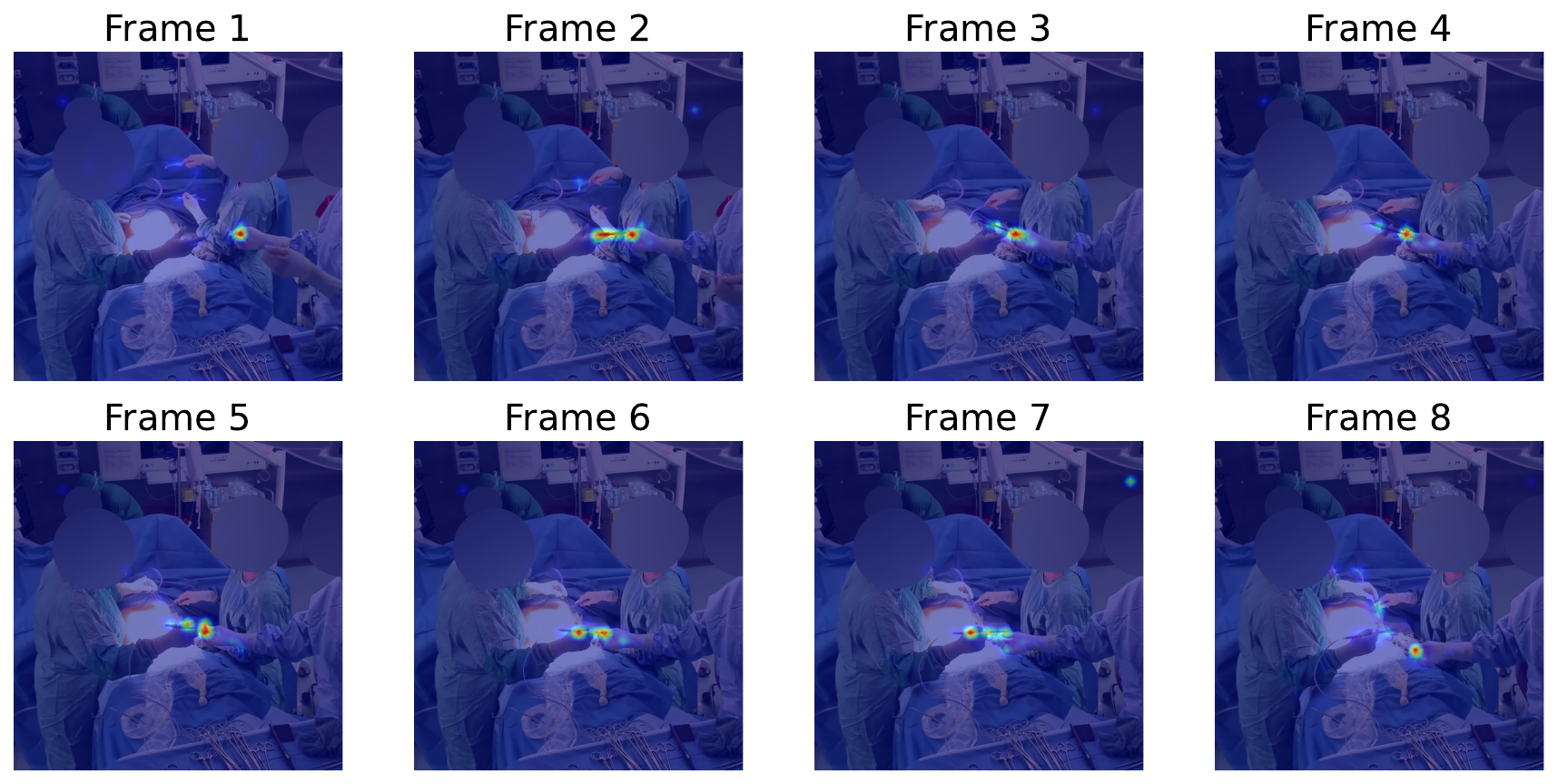}
        \caption{Training Set}
        \label{fig:layer-cam-whole-sequence-train}
    \end{subfigure}
    \hfill
    \begin{subfigure}{0.95\linewidth}
        \centering
        \includegraphics[width=\linewidth]{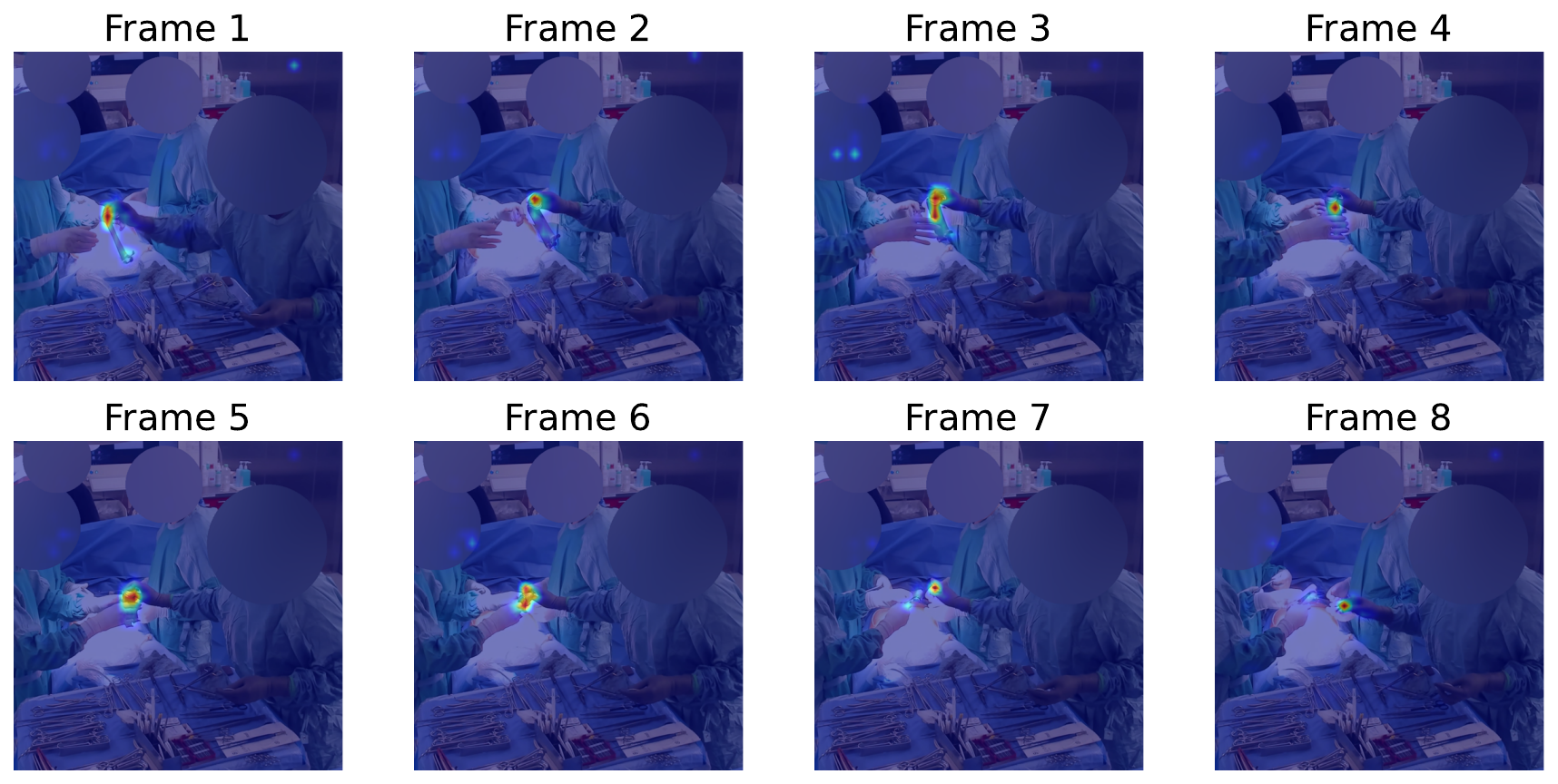}
        \caption{Test Set}
        \label{fig:layer-cam-whole-sequence-test}
    \end{subfigure}
    \caption{Layer-CAM explanation maps illustrating the contribution of each frame to the handover detection prediction. The explanation maps are generated using the second-to-last and third-to-last layers of the model.}
    \label{fig:layer-cam-whole-sequence}
\end{figure}

Additional qualitative examples of Layer-CAM explanations are provided in Fig.~\ref{fig:layer-cam-whole-sequence} to illustrate the spatial regions contributing to handover detection.

\end{document}